\documentclass[journal]{IEEEtran}
\IEEEoverridecommandlockouts

\usepackage{flushend}
\usepackage{cite}
\usepackage{amsmath,amssymb,amsfonts}
\usepackage{algorithmic}
\usepackage{graphicx}
\usepackage{textcomp}
\usepackage{xcolor}
\usepackage[colorlinks]{hyperref}
\usepackage[utf8]{inputenc} 
\usepackage[T1]{fontenc}    
\usepackage{hyperref}       
\usepackage{url}            
\usepackage{booktabs}       
\usepackage{amsfonts}       
\usepackage{nicefrac}       
\usepackage{microtype}      
\usepackage{multicol}
\usepackage{float}
\usepackage{multirow,array}
\usepackage{wrapfig}
\usepackage{lineno}
\usepackage{svg}
\usepackage[normalem]{ulem}
\usepackage{multicol}
\usepackage{color}
\usepackage{morefloats}

\usepackage{array}
\usepackage{booktabs}
\usepackage{multirow}
\usepackage{hyperref}
\usepackage{tkz-euclide}
\usepackage{tikz}
\usetikzlibrary{calc}
\usepackage{fp}
\usepackage{comment}
\usepackage{ marvosym }
\usepackage{caption}
\usepackage{subcaption}
\usepackage{authblk}
\usepackage{include/picins}
\usepackage{tikz}
\usepackage{comment}
\usetikzlibrary{shapes.geometric,arrows,chains,matrix,positioning,scopes,calc}
\tikzstyle{mybox} = [draw=white, rectangle]
\definecolor{darkblue}{rgb}{0,0.08,0.45}
\definecolor{blue}{rgb}{0,0,1}

\def\BibTeX{{\rm B\kern-.05em{\sc i\kern-.025em b}\kern-.08em
    T\kern-.1667em\lower.7ex\hbox{E}\kern-.125emX}}
\begin{document}

\title{Hyperbolic Generative Adversarial Network}

\author{Diego Lazcano,
        Nicol\'as Fredes,
        and Werner Creixell
\thanks{N. Fredes and D. Lazcano are master students at Department of Electronic Engineering, Universidad Técnica Federico Santa María, Valparaíso, Chile (e-mail: nicolas.fredes.13@sansano.usm.cl; diego.lazcano.13@sansano.usm.cl).}
\thanks{W. Creixell is Faculty at Department of Electronic Engineering, Universidad Técnica Federico Santa María, Valparaíso, Chile (e-mail: werner.creixell@usm.cl).}
}

\maketitle
\begin{abstract}
Recently, Hyperbolic Spaces in the context of Non-Euclidean Deep Learning have gained popularity because of their ability to represent hierarchical data. We propose that it is possible to take advantage of the hierarchical characteristic present in the images by using hyperbolic neural networks in a GAN architecture. In this study, different configurations using fully connected hyperbolic layers in the GAN, CGAN, and WGAN are tested, in what we call the HGAN, HCGAN, and HWGAN, respectively. The results are measured using the Inception Score (IS) and the Fr\'echet Inception Distance (FID) on the MNIST dataset. Depending on the configuration and space curvature, better results are achieved for each proposed hyperbolic versions than their euclidean counterpart.
\end{abstract}


\section{Introduction}

Deep learning architectures produce hierarchical representations, or abstract features, from the input data \cite{representation}. A vector or tensor represents the data, and then it is processed through multiple non-linear transformations called neural network layers \cite{visualizing_cnn}. The different layers types and their interconnection determine how the data is processed inside the network and consequently its representation. For each input sample, the representation or feature values depends on both the network structure and the network parameters. The neural layers are trained by optimizing their parameters to learn the input data patterns and manipulate their distribution through the network until obtaining an output with desirable statistical properties. The relationships and patterns, established by the networks, heavily rely on the assumption that the input data can be appropriately represented in a euclidean space. This assumption has been successful in settings dealing with images, sound, text, and other signals which clearly present an underlying euclidean structure. However, for data that present a non-euclidean nature, better performance can be expected using neural layers that exploit these geometrical properties.\newline

\begin{figure}[t]
    \centering
    \includegraphics[scale=0.6]{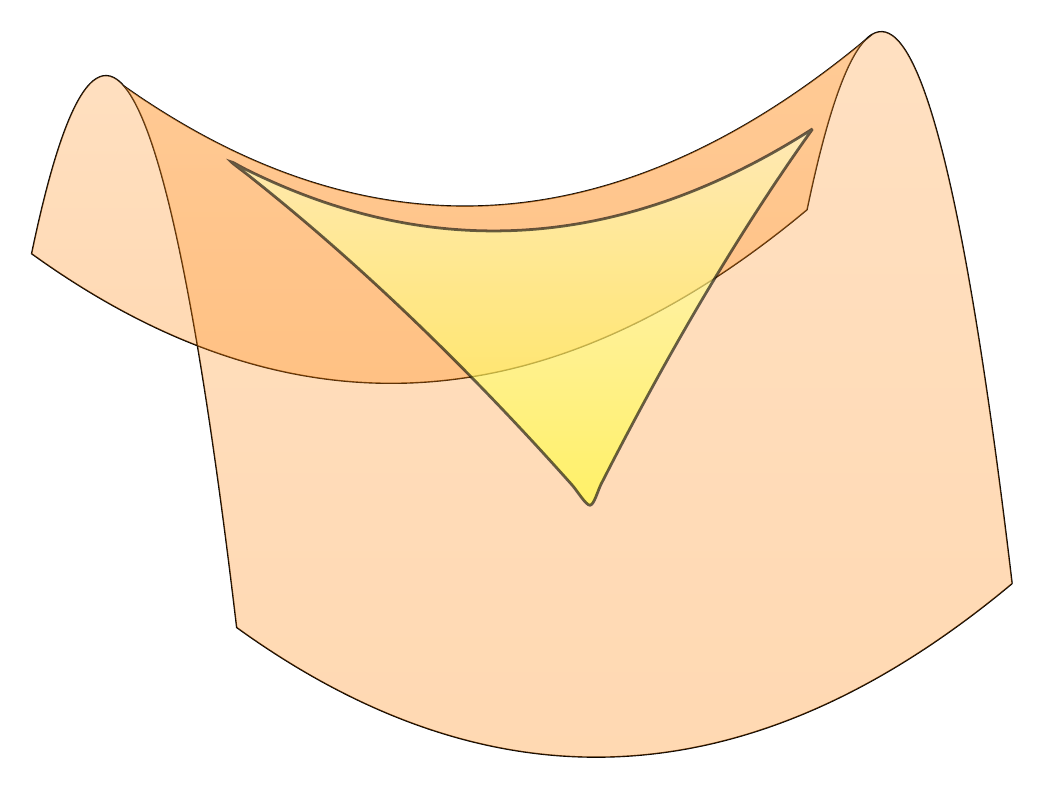}
    \caption{Triangle on a hyperbolic paraboloid.}
    \label{triangle_paraboloid}
\end{figure}

Geometric deep learning \cite{gdl}, also known as non-euclidean deep learning, is a novel approach to produce a better representation of non-euclidean data than conventional networks \cite{spherical}. In general, there are two kinds of non-euclidean data. The first are those that evidently can not be adequately represented by a euclidean space, like data in graphs. The second type is the data typically represented by euclidean geometry; however, they can present an underlying non-euclidean structure, such as cyclic or hierarchical relations. Hierarchy o cyclicity in the data can be better exploited  through embeddings in a Riemannian manifold. A Riemannian manifold is a ``curved'' space. Intuitively, the curvature concept can be understood by analyzing a surface. The surface curvature will be represented by how much any point in the surface deviates from a tangent plane. For spaces that present constant curvature, there are three families \cite{triangle}, those with positive curvature or elliptic spaces, negative curvature also known as hyperbolic spaces, and the Euclidean space with zero curvature \cite{non-euclidean-geometry}. In a Riemannian Manifold, the fifth axiom of Euclid (parallel postulate) does not hold \cite{rieman}, this postulate is equivalent to assert that in a triangle, the internal angles add 180° \cite{triangle}. Instead, depending on the intrinsic space curvature, the sum of the angles in a triangle can be bigger than 180° for elliptic spaces and less than 180° for hyperbolic spaces, as shown in figure \ref{triangle_paraboloid} for a triangle in a hyperbolic paraboloid.\newline

\begin{figure}[t]
\begin{center}
  \includegraphics[scale=0.4]{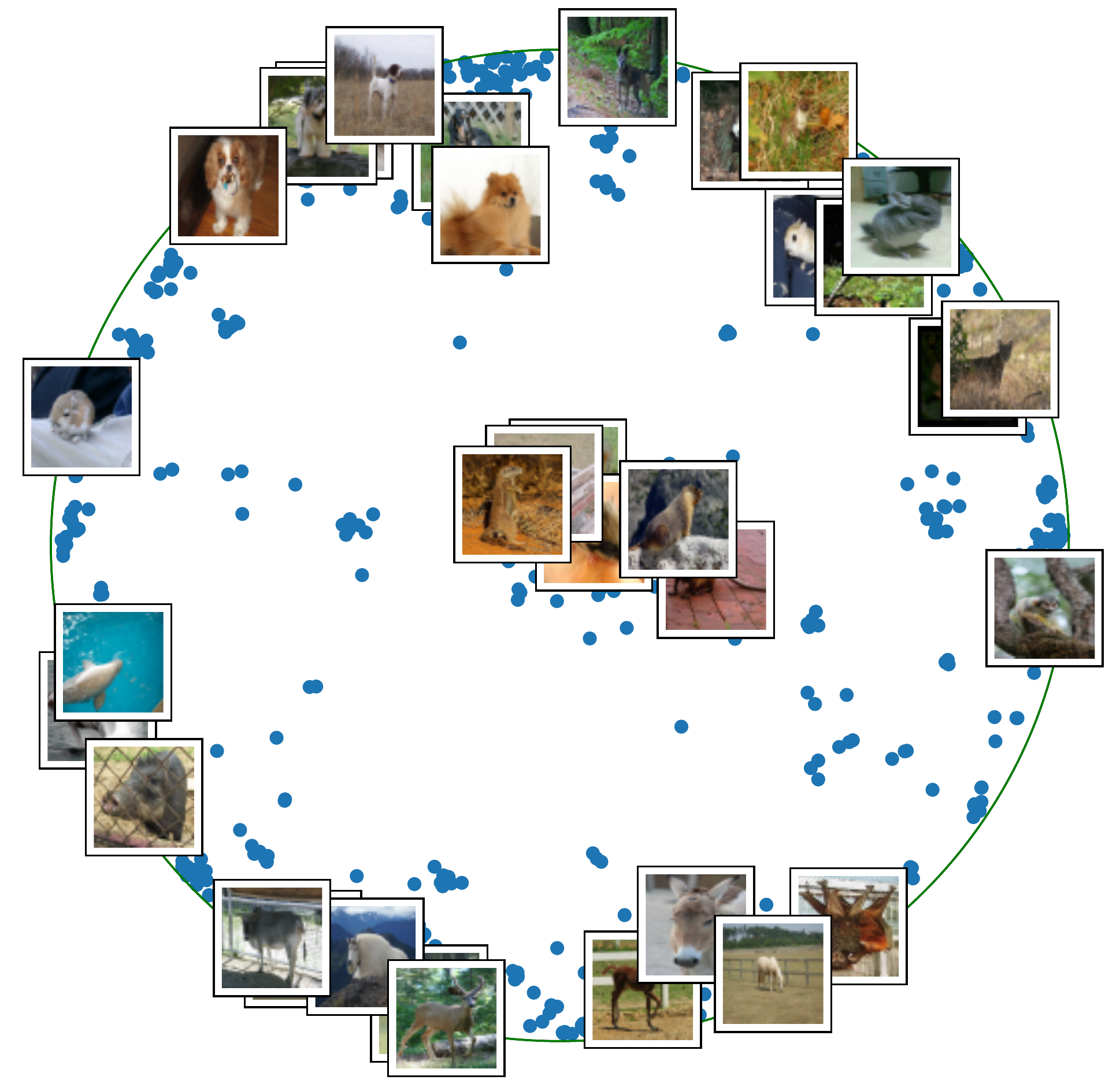}
  \caption{2D Poincaré embedding applied to WordNet database correlated with ImageNet dataset.}
  \label{poincare_embedding}
\end{center}
\end{figure}

Non-Euclidean spaces with constant curvature have properties useful for providing better representations for a certain type of structured data. The elliptical or spherical spaces are well suited for representing data with a cyclical structure \cite{mixedcurva} \cite{spherical_and_hyperbolic}. On the other hand, for hierarchically structured data, Hyperbolic spaces, particularly the Poincaré ball shares the same metric structure as trees \cite{hyp_geo_comp_net}. Hierarchical structural properties are strongly present in text and graphs \cite{hyp_link_pred}\cite{hyp_graph_cnn}, motivating different works using Poincaré Ball space. Tifreaa et al. proposed an adaptation of the GloVe \cite{glove} embedding algorithm in a Poincaré Ball \cite{poincglove} and Dhingra et al. developed a text embedding, using neural networks, in a hyperbolic space\cite{embedding_text_hyp_spaces}. In  \cite{poincemb}, a method is proposed to make link prediction over a graphs of words, embedded on a low dimension Poincaré ball, to achieve better results in word similarity and lexical entailment. However, besides text, other data types have underlying or latent tree structures, where Poincaré spaces have been less used. Images do not present an apparent hierarchical behavior; however, they can be grouped in classes because of their similarities. These classes can also be grouped in other classes, increasing the abstraction level or going up in a tree structure. An example of the hierarchy in images is the WordNet-ImageNet  \cite{wordnet}  \cite{imagenet} dataset. The WordNet dataset is a compound of words (nouns and adjectives) organized in a tree, and in the ImageNet dataset, the images are organized following the WordNet tree. The Hyperbolic space can leverage the hierarchy of ImageNet as shown in figure \ref{poincare_embedding}, where WordNet-ImageNet mammals tree embedded in a two-dimensional Poincaré Ball using the method proposed in \cite{poincemb}. This argument was first wielded in \cite{hyime} where they claim that the hierarchical semantic structure of language concepts can also be present in the images of those concepts, as in our example with mammals in figure \ref{poincare_embedding}. In that work, they modify the last section of three network architectures adding hyperbolic embedding to a Poincaré space followed by hyperbolic layers for the few-shot learning task. The hyperbolic layers were proposed in \cite{hnn} adapting the MLP, RNN, GRU, MLR to the hyperbolic space plus two layers for mapping from euclidean space to hyperbolic, called exponential map, and vice versa called logarithmic map. Hyperbolic Neural Networks (HNN) have been useful for the data exhibiting hierarchical latent anatomy because they increase its representation fidelity, with less distortion and dimensional requirements \cite{mixedcurva}.\newline

Similar to the embeddings, hyperbolic spaces have been employed to improve generation tasks on deep learning, both on text and images. Shuyang Dai et al. \cite{hyp_text_generation} developed text generation in hyperbolic space, using a hyperbolic version of a Variational Auto Encoder (VAE). Hyperbolic VAE's also have been used to hierarchically represent images on the Poincaré ball \cite{poinc_vae} \cite{wrapped_normal_hyp}. The VAE can be used to learn efficient representations and also for generation. However, the most popular architecture for the image generation tasks is the Generative Adversarial Network (GAN) \cite{gan}.\newline

GAN consists of two networks that participate in an adversarial game during training. This architecture comprises a generator network trying to generate data similar to the training dataset and a discriminator network trying to differentiate between real and generated data. The generator network is trained in an unsupervised way evaluating its output against the discriminator. The discriminator is trained in a supervised way for classifying real and fake images. After the networks have been trained, the generator can produce data similar to the dataset, but with original variations,  only from a noise input. There are multiple variations and modifications to the original GAN, but, to the best of our knowledge, there is no application of hyperbolic space to GANs architectures that aim to exploit the data's hierarchical characteristics, as in the VAE architecture.\newline

We propose that the GAN architecture can take advantage of the hierarchical characteristics present in the images by using hyperbolic neural networks. Thereby, they can achieve better quality and diversity of the generated images. In this work, we use the GAN \cite{gan}, WGAN \cite{wgan}, and CGAN \cite{cgan} architectures, conforming to their hyperbolic versions HGAN, HWGAN, and HCGAN. In each case, experiments with different arrangements and combinations of euclidean with hyperbolic layers and different curvature values were conducted. The performance over the MNIST dataset was measured by the Inception Score (IS) \cite{inception_score}, and Fréchet Inception Distance (FID) \cite{fid}, achieving better results for some layers arrangements for each architecture.

\section{Background}
\subsection{The Poincaré Ball}

The hyperbolic spaces $\mathbb{H}^n$ are $n$-dimensional Riemannian manifolds homogeneous and simply connected with a constant negative curvature. There are multiple models of hyperbolic space, but this research focuses on The Poincaré Ball. The Poincaré Ball manifold $(\mathbb{D}^{n}_c, g^{\mathbb{D}^n_c}_\mathbf{u})$  is a $n$-ball over $\mathbb{R}^n$ of radius $1/\sqrt{c}$. In this space, the ball center is the hyperbolic version of euclidean zero, and the ball perimeter is the infinity.\newline\newline
The Poincaré Ball $(\mathbb{D}^{n}_c, g^{\mathbb{D}^n_c}_\mathbf{u})$  is defined as:
\begin{equation}\label{eq:poincare1}
    \mathbb{D}_{c}^{n}=\{\mathbf{u}\in \mathbb{R}^{n}: c\:||\mathbf{u}||^{2}<1,\:c \geq 0\}
\end{equation}
With the Riemannian metric tensor:
\begin{equation}\label{eq:poincare3}
    g_{\mathbf{u}}^{\mathbb{D}^n_{c}} = \lambda_{\mathbf{u}}^{c}\:g^{E}
\end{equation}
\begin{equation}
    \lambda_{\mathbf{u}}^{c} := \frac{2}{1-c\:||\mathbf{u}||^{2}}
\end{equation}
$\lambda_{\mathbf{u}}^c$ is called the conformal factor of this space, and $g^{E}=I_{n}$ is the Euclidean metric tensor in the cannonical base.\newline
With $\mathcal{T}_{\mathbf{u}}$ the tangent space operator, the exponential map takes a vector $\mathbf{x} \in \mathcal{T}_{\mathbf{u}}\mathbb{D}_{c}^{n} \cong \mathbb{R}^{n}$ and assign it to a vector $\mathbf{v} \in \mathbb{D}_{c}^{n}$. Its equation is given by:
\begin{equation}
\label{expx}
    \exp_{\mathbf{u}}^{c}(\mathbf{x}):=(\mathbf{u}) \oplus_{c} \left(\tanh\left(\sqrt{c}\;\frac{||\mathbf{x}||\lambda_{u}^{c}}{2}\right)\frac{\mathbf{x}}{\sqrt{c}||\mathbf{x}||}\right) \ 
\end{equation}
where the Möbius addition $\oplus_{c}$, for $\mathbf{u}, \mathbf{v} \in \mathbb{D}^{n}_c$, is defined as:
\begin{equation}\label{moebius_add}
\mathbf{u} \oplus_{c} \mathbf{v}:=\frac{\left(1+2 c\langle\mathbf{u}, \mathbf{v}\rangle+c\|\mathbf{v}\|^{2}\right) \mathbf{u}+\left(1-c\|\mathbf{u}\|^{2}\right) \mathbf{v}}{1+2 c\langle\mathbf{u}, \mathbf{v}\rangle+c^{2}\|\mathbf{u}\|^{2}\|\mathbf{v}\|^{2}}
\end{equation}
 The inverse of exponential map, the logarithmic map, takes a vector $\mathbf{v} \in \mathbb{D}_{c}^{n}$ and assign it to a vector $\mathbf{x} \in \mathcal{T}_{u}\mathbb{D}_{c}^{n} \cong \mathbb{R}^{n}$.
 \begin{equation}
 \label{logx}
    \log_{\mathbf{u}}^{c}(\mathbf{v}):=\frac{2}{\sqrt{c}\:\lambda_{\mathbf{u}}^{c}}\tanh^{-1}(\sqrt{c}||-\mathbf{u} \oplus_{c}\:\mathbf{v}||)\frac{-\mathbf{u}\oplus_{c}\mathbf{v}}{||-\mathbf{u}\oplus_{c} \mathbf{v}|| }
 \end{equation}
 
For simplicity we choose $\mathbf{u}$ as the origin of $\mathbb{R}^{n}$ and the equations \eqref{expx} and \eqref{logx} result in:
 \begin{equation}
 \label{exp0}
    \mathbf{v} = \exp_{0}^{c}(\mathbf{x}):=\tanh(\sqrt{c}||\mathbf{x}||)\frac{\mathbf{x}}{\sqrt{c}||\mathbf{x}||}
\end{equation}
 
 \begin{equation}
 \label{log0}
     \mathbf{x} = \log_{0}^{c}(\mathbf{v}) := \tanh^{-1}(\sqrt{c}||\mathbf{v}||)\frac{\mathbf{v}}{\sqrt{c}||\mathbf{v}||}
 \end{equation}


\subsection{Hyperbolic Deep Learning}

\begin{figure}[t]
\begin{center}
  \includegraphics[scale=0.65]{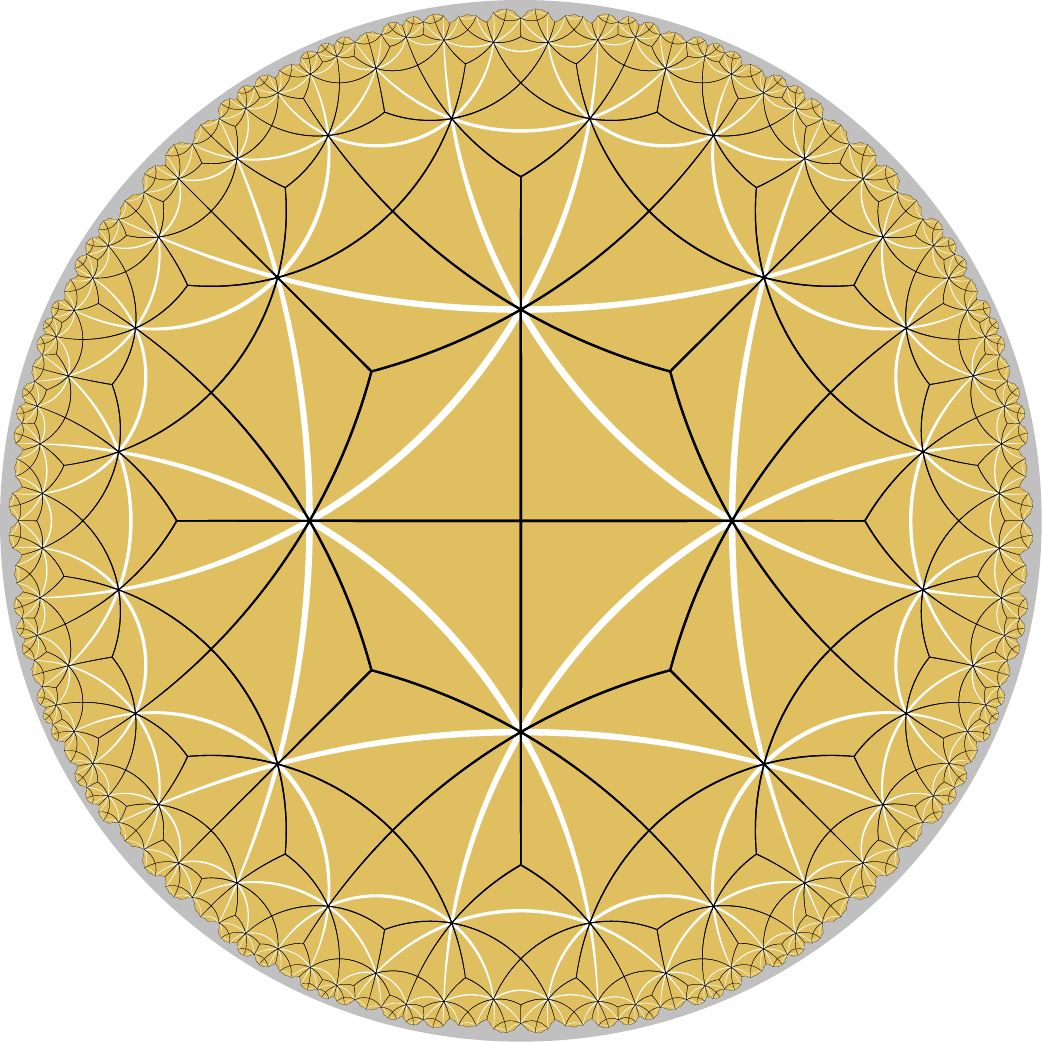}
  \caption{Tesselation of squares (sides of equal length) in a 2D Poincar\'e ball. The squares sides are in black, and the squares diagonals are in white.}
  \label{poincare_ball}
\end{center}
\end{figure}

Hyperbolic spaces have a high capacity to represent data with tree-like structures. Figure \ref{poincare_ball} shows how a square grid shapes into a tree like-structure in a 2D Poincaré ball. When applied to neural network layers, this property allows a better representation of hierarchical data, which is exploited in Hyperbolic Neural Networks (HNN's) \cite{hnn}. The HNN uses the gyrovectors to implement the basic neural network operations in the Poincaré Ball. The gyrovector spaces allows create a simil of a vector space in a non-eculidean space like the hyperbolic. The gyrovector space used to operate in the Poincaré Ball is the Möbius, with two binary operations the Möbius addition \eqref{moebius_add}, and the Möbius scalar multiplication of $k\in \mathbb{R}$ by vector $\mathbf{v} \in \mathbb{D}^{n}_{c}$, defined as:
\begin{equation}
\label{eq:escalar_vector}
    k\otimes_{c} \mathbf{v}:= \exp_{0}^{c}\left(r \cdot \log_{0}^{c}(\mathbf{v}) \right)
\end{equation}

 Equation \eqref{eq:escalar_vector} together with equation \eqref{moebius_add} allow us to build feedfordward networks. 
A Feedforward network consists of an affine transformation and a non-linear function as activation. The hyperbolic feedforward network \cite{hnn}
has a gyrovector-based affine transformation with Möbius matrix-vector multiplication and Möbius addition the bias vector. The Möbius matrix-vector multiplication was defined in \cite{hnn}; the procedure implementation is similar to Möbius scalar multiplication \eqref{eq:escalar_vector}, it uses the exponential and logarithmic mappings. For $ \mathbf{v} \in \mathbb{D}^{n}_{c} $ to be multiplied by a Matrix $M \in \mathbb{R}^{m\times n}$ , $\mathbf{v}$ is taken to the Euclidean space by a logarithmic map, and  multiplied by $M$. The exponential map is then applied to the resulting vector. This operation  ($\mathbb{D}^{n}_c\rightarrow \mathbb{D}^{m}_c$) is given by:

\begin{equation}
    M\otimes_{c}\mathbf{v}:= \exp_{0}^{c}\left(M \cdot \log_{0}^{c}(\mathbf{v}) \right)
\end{equation}

The bias Möbius addition is a translation of gyrovector $\mathbf{v} \in \mathbb{D}^{n}_{c}$ by bias $\mathbf{b} \in \mathbb{D}^{n}_{c}$, that is given by:

\begin{equation}
     \mathbf{v} \oplus_{c} \mathbf{b} = \exp_{\mathbf{v}}^{c}\left(\frac{\lambda_{0}^{c}}{\lambda_{\mathbf{v}}^{c}}\log_{0}^{c}(\mathbf{b})\right),
\end{equation}

Additionally, in order to apply any function to a gyrovector, the Möbius version of the function is required. Similarly to Möbius matrix-vector multiplication, let $f:\mathbb{R}^{n}\rightarrow \mathbb{R}^{m}$, the Möbius version $f^{\otimes_{c}}$ that map from $\mathbb{D}^{n}_{c}$ to $\mathbb{D}^{m}_{c}$ is defined by:

\begin{equation}
    f^{\otimes_{c}}(\mathbf{v}):=\exp_{0}^{c}(f(\log_{0}^{c}(\mathbf{v})))
\end{equation}

\subsection{GAN}
Generative Adversarial Networks \cite{gan} are an architecture of two networks that participate in an adversarial game during training. One, the Generator, is in charge of producing artificial images from noise input. The other, the discriminator, is in charge of classifying between real images from the artificially generated ones. Therefore, during training, the generated images are passed to the discriminator input, which in turn alternates artificial and real images. The roles in the adversarial game are as follows: the generator $\mathcal{G}$ is trying to produce images similar to the real ones in order to fool the discriminator $\mathcal{D}$; on the other hand, the discriminator is trying to detect whether a particular image is real or artificially generated by $\mathcal{G}$. In this way, the generator $\mathcal{G}(\mathbf{z};\theta_g)$ learn the distribution $p_{data}$ of the images $\mathbf{x}$ from the noise input $\mathbf{z}$, and the discriminator $\mathcal{D}(\mathbf{x};\theta_d)$ estimates the probability of $\mathbf{x}$ being real or artificial. Equation \eqref{gan} shows the loss function for the GAN, a minmax game, as proposed by \cite{gan}.

\begin{equation}
 \label{gan}
    \min \limits_{\mathcal{G}} \max \limits_{\mathcal{D}} V(\mathcal{G}, \mathcal{D}),
\end{equation}

Where

\begin{align}
    V(\mathcal{G}, \mathcal{D})=&\mathbb{E}_{\mathbf{x} \sim p_{data}(\mathbf{x})}[\log (\mathcal{D}(\mathbf{x}))] \nonumber \\ 
    & + \mathbb{E}_{\mathbf{z}\sim p_{z}(\mathbf{z})}[\log (1-\mathcal{D}(\mathcal{G}(\mathbf{z})))]
\end{align}

The GAN's training finishes when the discriminator is unable to distinguish between real images from generated images. Once trained, the generator network can be used independently for generating images. However, it is impossible to control the output in any way. Furthermore, the GAN network can have challenges to train because of vanishing gradients, and mode collapse; more details about these issues can be found in \cite{gan_train}. In this work, in addtition to the GAN,  we use two modifications to the original GAN architecture, the  Conditional GAN  (CGAN) \cite{cgan}, and the Wasserstein GAN (WGAN) \cite{wgan}.\newline

The CGAN has additional class label information concatenated in the input layer of the generator and discriminator network. This additional information condition both networks, like a digit label, when working with the MNIST dataset.  Equation \eqref{cgan} shows the loss function for the CGAN, which includes the label information $\mathbf{y}$ for conditioning at the discriminator and generator.


\begin{align}
\label{cgan}
    V(\mathcal{G}, \mathcal{D})=&\mathbb{E}_{\mathbf{x} \sim p_{data}(\mathbf{x})}[\log (\mathcal{D}(\mathbf{x}|\mathbf{y}))] \nonumber \\ 
    & + \mathbb{E}_{\mathbf{z}\sim p_{z}(\mathbf{z})}[\log (1-\mathcal{D}(\mathcal{G}(\mathbf{z}|\mathbf{y})|\mathbf{y}))]
\end{align}

The WGAN addresses the issues of vanishing gradients and mode collapse of the original GAN architecture. To achieve this goal, they use Wasserstein distance to have a smoother gradient everywhere, but the distance equation is intractable. The Kantorovich-Rubinstein duality simplifies the equation of distance as \eqref{wgan}:

\begin{equation}\label{wgan}
W\left(\mathbb{P}_{r}, \mathbb{P}_{\theta}\right)=\sup _{\|f\|_{L} \leq 1} \mathbb{E}_{\mathbf{x} \sim \mathbb{P}_{r}}[f(\mathbf{x})]-\mathbb{E}_{\mathbf{x} \sim \mathbb{P}_{\theta}}[f(\mathbf{x})]
\end{equation}

 Where $\sup$ indicate the supremum, $\mathbb{P}_{r}$ and $\mathbb{P}_{\theta}$ two distributions, and $f$ is a 1-Lipschitz function that meet the following constraint:
 
 \begin{equation}
     |f(x_{1})-f(x_{2})|\leq |x_{1}-x_{2}|.
 \end{equation}
 
  To calculate the Wasseterin distance is necessary to find a 1-Lipschitz  function. However, in practice this function can be learned by a neural network, and the discriminator network $\mathcal{D}$, without sigmoid function in the output layer, is an ideal candidate for doing this task. Under this configuration, the output of the discriminator can be any real number, and the bigger this score, the closest to a real image the input image is. To enforce $\mathcal{D}$ to comply with 1-Lipschitz restrictions, WGAN can apply a gradient penalty \cite{wgan-gp}. The WGAN with Gradient Penalty (WGAN-GP) works since any differentiable function $f$ is 1-Lipschitz if and only if its gradient have norm less or equal to one in the whole space (check the demonstration in \cite{wgan-gp}). The loss function of the WGAN-GP is given by:

\begin{align}
\min \limits_{\mathcal{D}} & \mathbb{E}_{\mathbf{z}\sim p_{z}(\mathbf{z})}[\mathcal{D}(\mathcal{G}(\mathbf{z}))] \: - \:  \mathbb{E}_{\mathbf{x}\sim p_{data}(\mathbf{x})}[\mathcal{D}(\mathbf{x})]\nonumber \\
&+\lambda \: \mathbb{E}_{\mathbf{\hat{x}}\sim p_{\hat{x}}(\mathbf{\hat{x}})}[\left(||\nabla_{\mathbf{\hat{x}}}\mathcal{D}(\mathbf{\hat{x}})  ||_{2}-1\right)^{2}]
\end{align}

\begin{equation}
    \max \limits_{\mathcal{G}}  \mathbb{E}_{\mathbf{z}\sim p_{z}(\mathbf{z})}[\mathcal{D}(\mathcal{G}(\mathbf{z}))] 
\end{equation}

Where $\mathbf{\hat{x}}$ sampled from fake images $\mathcal{G}(\mathbf{z})$, and real images $\mathbf{x}$ with $\epsilon$ uniformly sampled between 0 and 1.

\begin{equation}
    \mathbf{\hat{x}} = \epsilon \: \mathbf{x} + (1- \epsilon)\:\mathcal{G}(\mathbf{z})
\end{equation}

\section{Hyperbolic GAN (HGAN)}
\begin{figure*}[t]
    \centering
    \subcaptionbox{Generator Network \label{fig:generator}}
     {\includegraphics[scale=0.35]{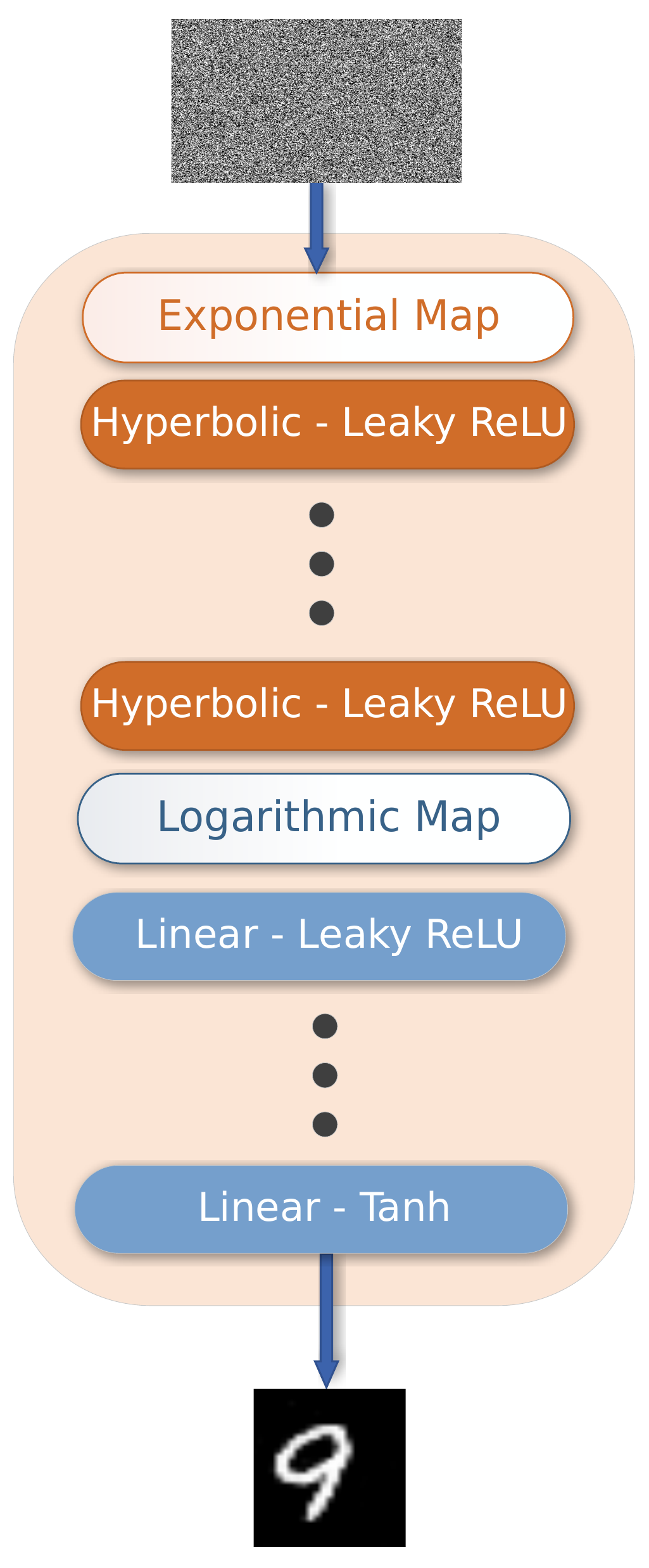}}
    \subcaptionbox{Discriminator Network \label{fig:discriminator}}
        {\includegraphics[scale=0.35]{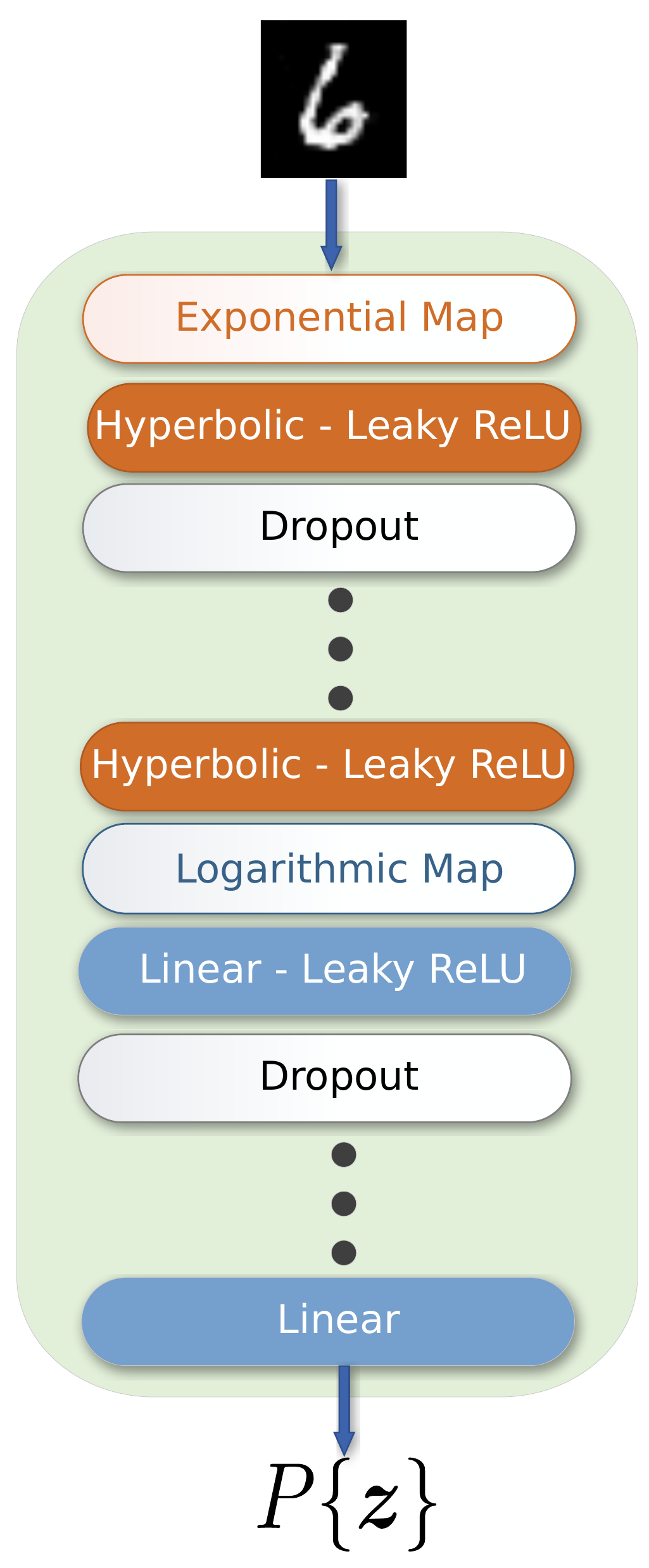}}
\caption{HGAN Architecture}
\label{fig:arch}
\end{figure*}

The central assumption in this work is that the hyperbolic neural networks can improve GAN's performance, either in the generation or discrimination process, because the hyperbolic layers can leverage the hierarchical characteristics of the images\cite{hyime}. This hyperbolic version of GAN, the HGAN, mix the hyperbolic and euclidean space in either the generator and discriminator, replacing some of the euclidean linear layers by the hyperbolic linear layers implemented in \cite{hyime}; therefore, the use of exponential and logarithmic mapping is necessary to move between euclidean and hyperbolic spaces. The use of different spaces in the neural networks allows the creation of abstract features representing different implicit data structures, like the hierarchical structure as already mentioned. This approach adds more degree of freedom to the HGAN design by the number of hyperbolic layers, their location, and the hyperbolic space curvature (represented by $c$). The HGAN is a family of architectures derived from modifying the original GAN architecture proposed by Goodfellow et al. \cite{gan}, figure \ref{fig:arch} shows the general architecture that we implemented. Similarly to the HGAN, the HWGAN, and HCGAN, correspond to modifications of the original WGAN-GP \cite{wgan-gp}, and CGAN \cite{cgan} architectures by replacing some euclidean by hyperbolic layers.\newline

The HGAN family architectures can have both euclidean layers and hyperbolic layers in different arrangements, with the corresponding exponential and logarithm mapping, center on zero, necessary to pass from the euclidean space to hyperbolic space and vice versa. The notation of this architecture's different configurations are $\mathcal{D}_{eehh}$, for the discriminator, and $\mathcal{G}_{hhee}$, for the generator, the subscripts denote which layers they are composed of. For example, $\mathcal{G}_{hhee}$ means that the generator network comprises an exponential map, two hyperbolic layers, a logarithmic map, and two euclidean layers,  the reading order from left to right correspond the order from input to output in the network. For simplicity, both exponential and logarithmic mappings are omitted from the notation. With this notation, the GAN architecture is represented by $\mathcal{D}_{eeee}\mathcal{G}_{eeee}$. \newline

The place of hyperbolic layers in the network has a significant influence on the HGAN performance. Three different general configurations were tested. First, the HGAN with a euclidean-hyperbolic ($EH$) configuration, consisting of euclidean layers, an exponential map, and finalize with hyperbolic layers. The $EH$ configuration first generates a feature vector in euclidean space, and then the abstract representations are processed in the Poincaré Ball. We believe that the $EH$ configurations could improve the discriminator network's performance because the image hierarchical structure should be relevant in deep process layers where there is a more abstract representation. The second configuration is the hyperbolic-euclidean ($HE$), which consists of the exponential map in the network input, followed by hyperbolic layers, logarithmic map, and euclidean layers in the network output. This sequence, $HE$, could improve performance by creating hierarchical representations first in the hyperbolic space that can help the euclidean layers task. Finally, we have the Euclidean-Hyperbolic-Euclidean ($EHE$) configurations that consist of the network start with euclidean layers, followed by the exponential map, hyperbolic layers, logarithmic map, and euclidean layers again to the network end. The $EHE$ configuration first creates a representation in the euclidean space enriched with abstract features that can then be exploited by the hyperbolic layers, and the final layers map it to euclidean space. The generator can better exploit this characteristic since it has to create an entire image from a complete non hierarchical noise input. \newline

 \begin{figure*}[h]
\begin{center}
  \centering
  \includegraphics[scale=0.6]{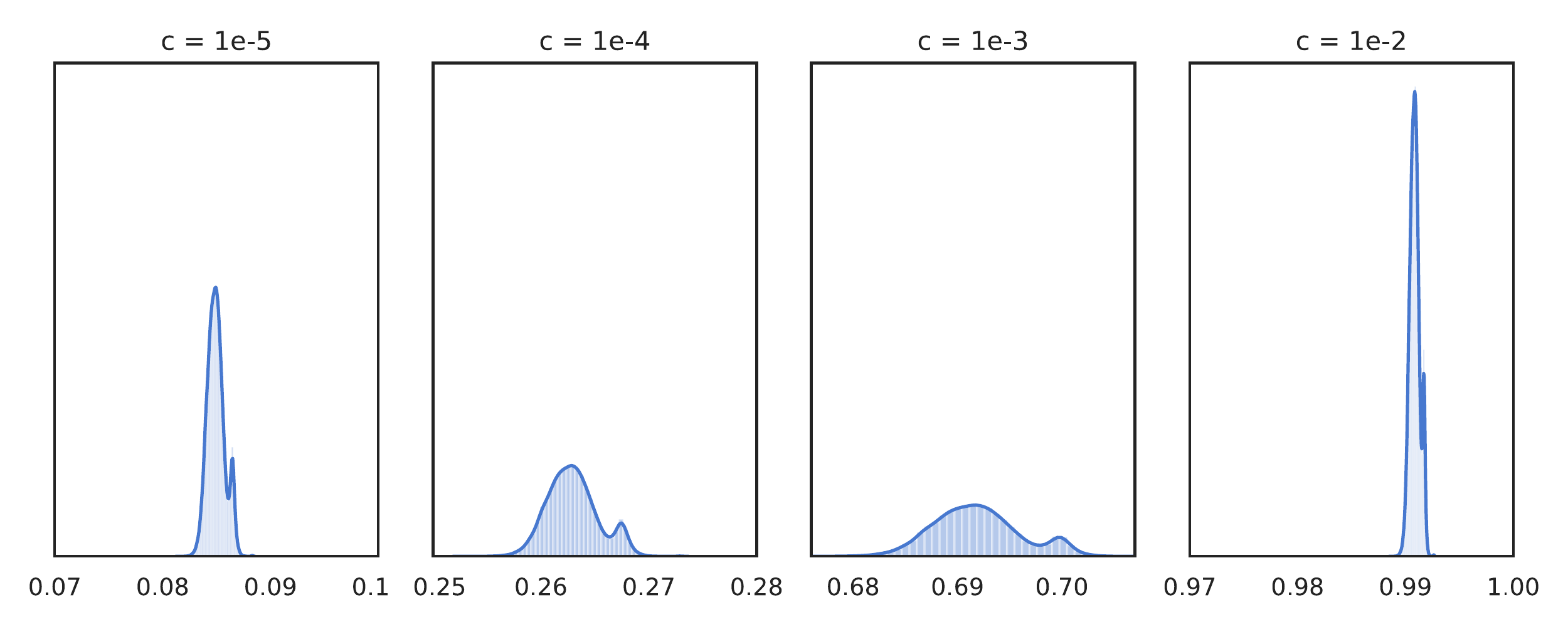}
  \caption{Poincaré Ball embedded vectorized MNIST ($\mathbf{x}$) distribution, from equation \eqref{exp0} normalized respect the Poincaré ball radius $r$ using the function $R(\mathbf{x}) = ||\exp_{0}^{c}(\mathbf{x})||/r = \tanh(\sqrt{c}||\mathbf{x}||)$ .}
  \label{fig:mnist_embedded}
\end{center}
\end{figure*}

Another degree of freedom is the $c$ value, which has a significant influence on HGAN performance. The Poincaré Ball radius $r$ is related to $c$ by $r = 1/\sqrt{c}$. Therefore, $c$ represents how big the Poincaré ball is, and its effect when mapping the MINIST dataset can be seen in figure \ref{fig:mnist_embedded}. The distribution of the magnitude of the images in the hyperbolic space change as $c$ varies, for $c=10^{-5}$ ($r\approx 316$) the distribution is near 0; on the other hand, for $c=10^{-2}$ ($r=10$) the distribution is squashed to the $r$ value. A wrong selection of $c$ causes a poor behavior or fault in the convergence because the data can collapse into the ball's boundary or to the origin and it can also produce numeric instability. The study of the appropriate $c$ values for a specific data set is a crucial task.\newline

Finally, the HGANs architectures used Leaky ReLu directly in hyperbolic space as an activation function, without a logarithmic and exponential map, which differs from the standard procedure to implement functions in the hyperbolic space. Furthermore, the HGAN has standard dropout layers for regularization, and the method used for optimization was Adam \cite{adam}. 



\section{Experiments}


The experiments consist of training the architectures with the MNIST dataset. The performance of each architecture is measure with the Inception Score (IS) \cite{is_mnist} and the Fr\'echet Inception Distance (FID) \cite{fid}. \newline


The GAN implementation has four linear layers on each network. The discriminator input is a vectorized image with 784 dimensions, followed by layers of 1024, 512, 256, and 1 hidden units respectively. For the generator, the input layer has  128 units, which corresponds to Gaussian noise, and the hidden layers are of size 256, 512, 1024, and 784, respectively. The discriminator contains dropout layers with a rate of 0.1. Both networks have a leaky ReLU with a leak factor equal a 0.2 for activation, except the network end, where the discriminator has no activation function, and the generator has an hyperbolic tangent. The loss function was the binary cross-entropy with logistic regression, and the training pipeline used the Adam \cite{adam} optimizer with a learning rate equal to 0.001, $\beta_{1}=0.5$, $\beta_{2}=0.999$. The WGAN-GP used the same structure as the GAN. However, it uses the Wasserstein loss with gradient penalty \cite{wgan} and Adam optimizer with a learning rate equal to 0.0001 and the same beta values. The version of CGAN follows almost the same structure, but with the discriminator input of 794 and the generator input of 138, ten more nodes on both network in order to allocate the one-hot encoding for the class labels. The optimizer uses the same set of parameters than the WGAN-GP with binary cross entropy with logistic regression as loss function.\newline

\begin{figure*}[h]
    \centering
    \includegraphics[scale=0.45]{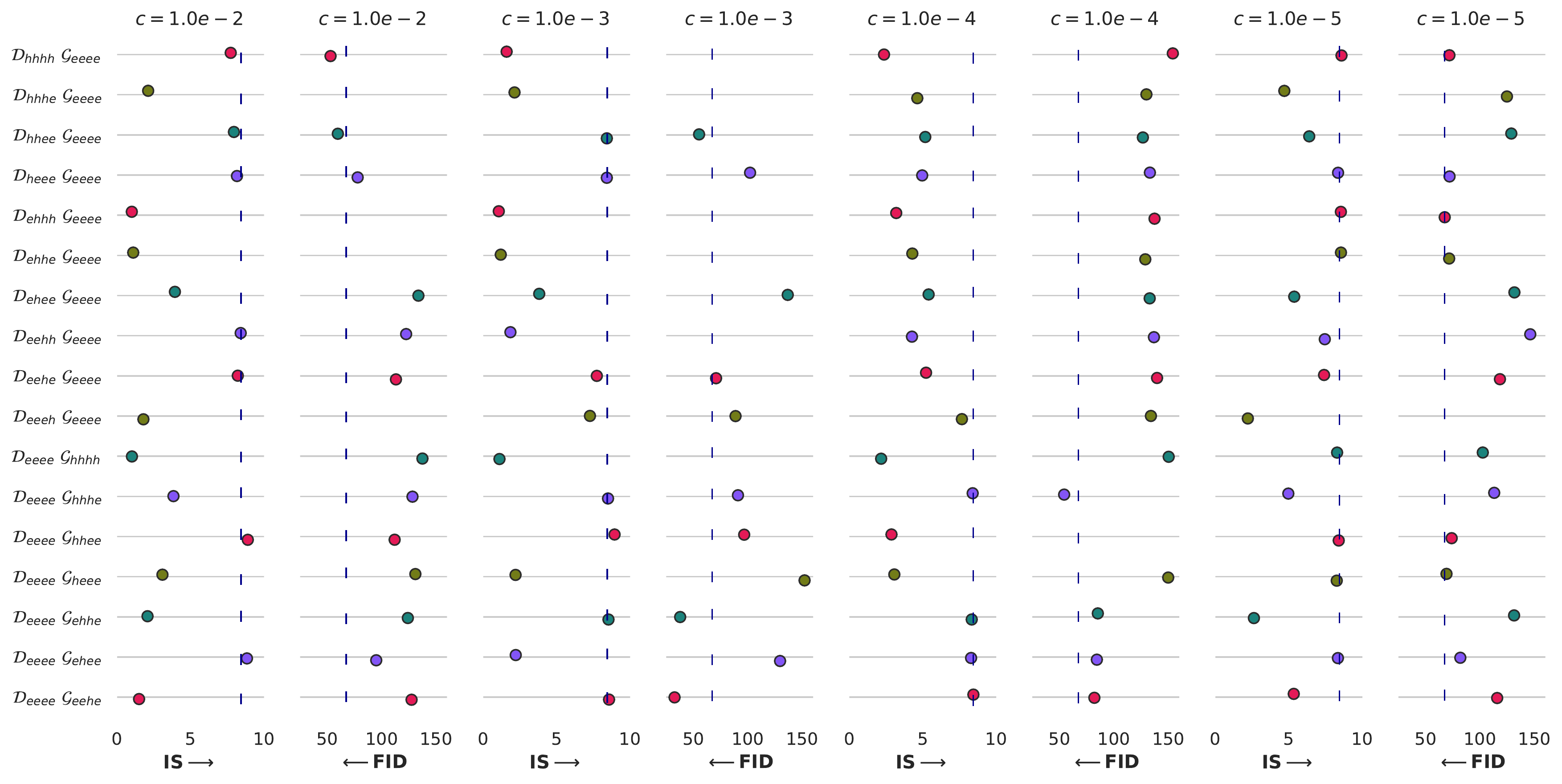}
    \caption{HGAN experiments visualization with only hyperbolic discriminator or hyperbolic generator measure  with  FID and IS. The segment line indicate the euclidean GAN performance.}
    \label{fig:hgan_table}
\end{figure*}

 The tested networks were compound of blocks of hyperbolic networks at the begging ($HE$), in the middle ($EHE$), and to the end ($EH$) for both generator and discriminator with different fixed $c$ values. For configurations where both discriminator and generator have hyperbolic linear layers, combinations of the architectures in the previous step were combined into one, with the sole exception of HWGAN, where the discriminator never showed better performance than the euclidean version. Since the different nature of discriminator and generator tasks, the curvature for each was not always the same, $c_d$ was for the discriminator and $c_g$ for the generator. This distinction produced a remarkable improvement in the FID score from 67.291 for the euclidean GAN to 18.697 for the HGAN with $c_d=10^{-5}$ and $c_g=10^{-3}$, as showed in table \ref{table_hgan}. The best performing architectures are summarized in tables \ref{table_hgan} for the HGAN, \ref{table_hwgan} for the HWGAN, and \ref{table_hcgan} HCGAN. The broader exploratory study obtained in this paper are shown in the annex figures \ref{fig:hgan_table}, \ref{fig:hcgan_table}, \ref{fig:hwgan_table_d}, and \ref{fig:hwgan_table_g}.

\begin{figure*}[t]
    \centering
\begin{tabular}{c c c}
     \begin{subfigure}[b]{0.25\textwidth}
         \centering
         \includegraphics[width=\textwidth]{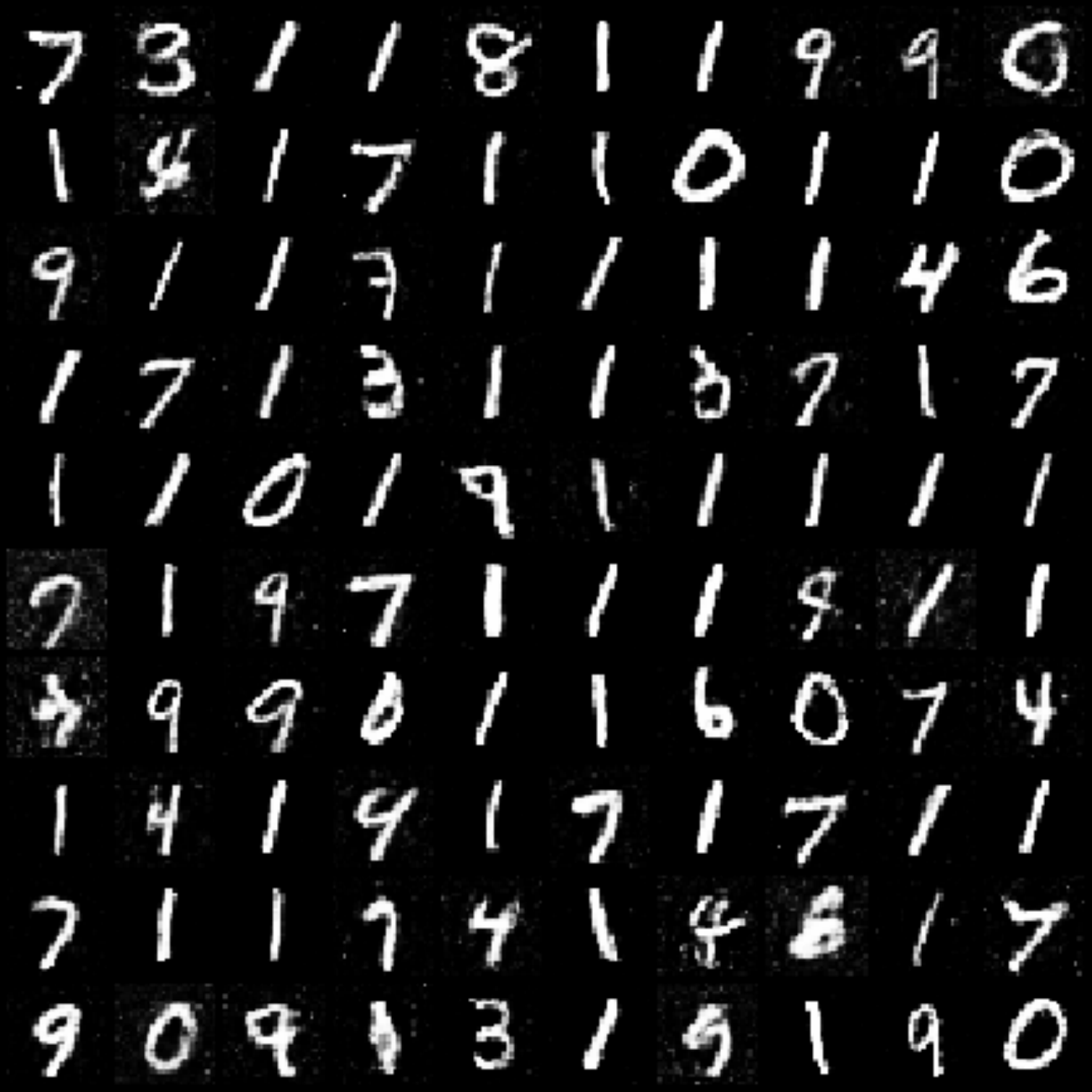}
         \caption{GAN}
         \label{fig:exp:mnist_gan}
     \end{subfigure}&
      \begin{subfigure}[b]{0.25\textwidth}
         \centering
         \includegraphics[width=\textwidth]{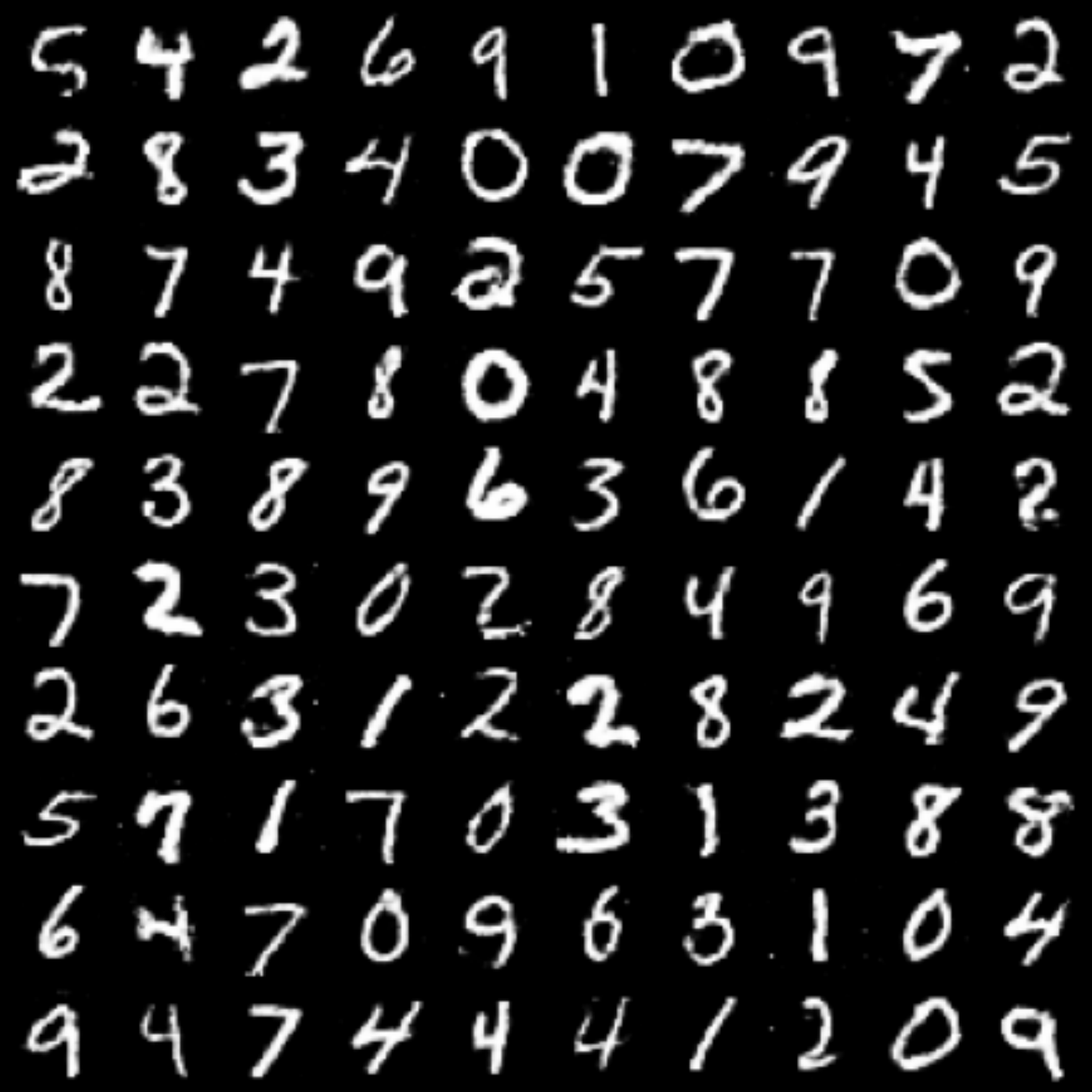}
         \caption{CGAN}
         \label{fig:three sin x}
     \end{subfigure}&
     \begin{subfigure}[b]{0.25\textwidth}
         \centering
         \includegraphics[width=\textwidth]{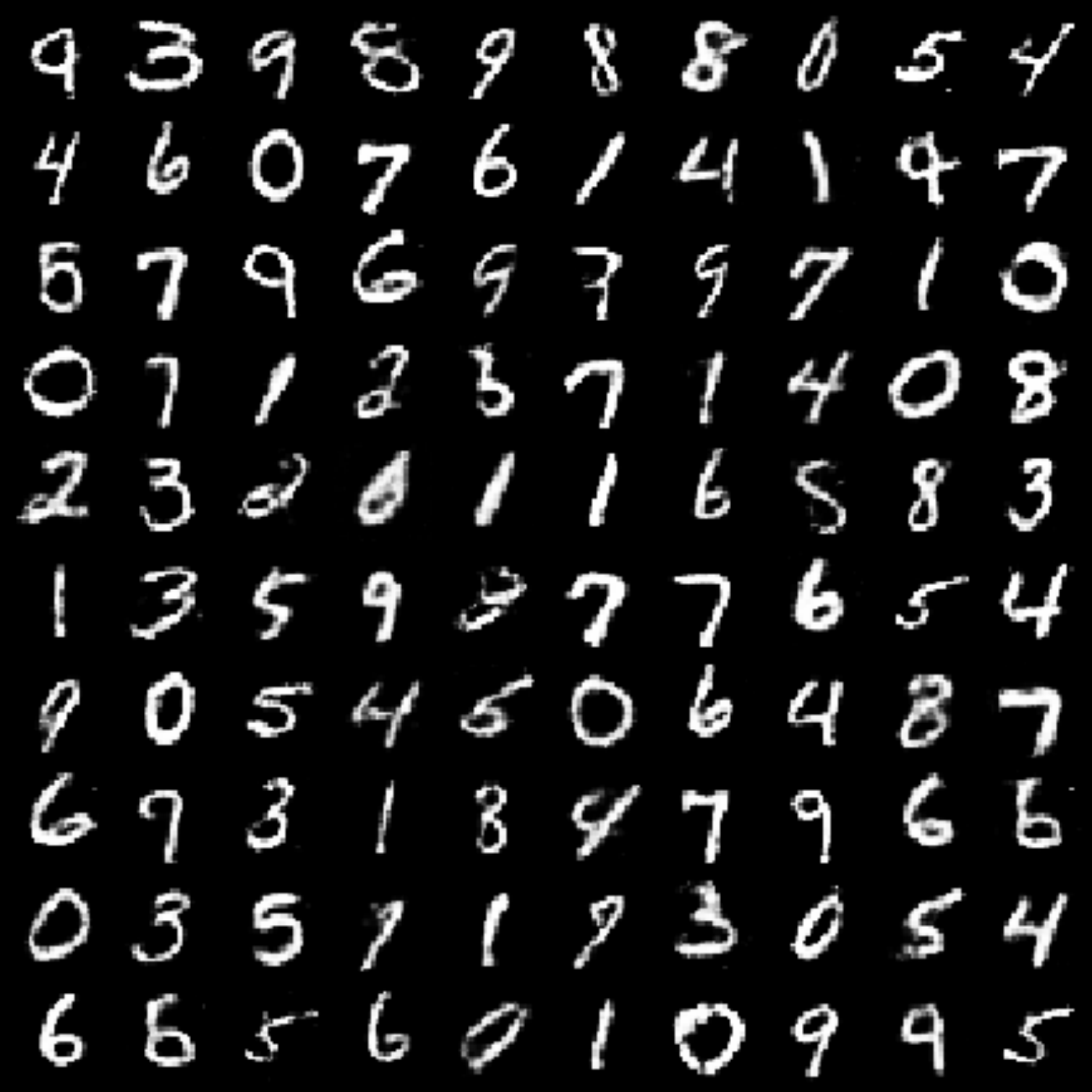}
         \caption{WGAN}
         \label{fig:three sin x}
     \end{subfigure}\\
     \begin{subfigure}[b]{0.25\textwidth}
         \centering
         \includegraphics[width=\textwidth]{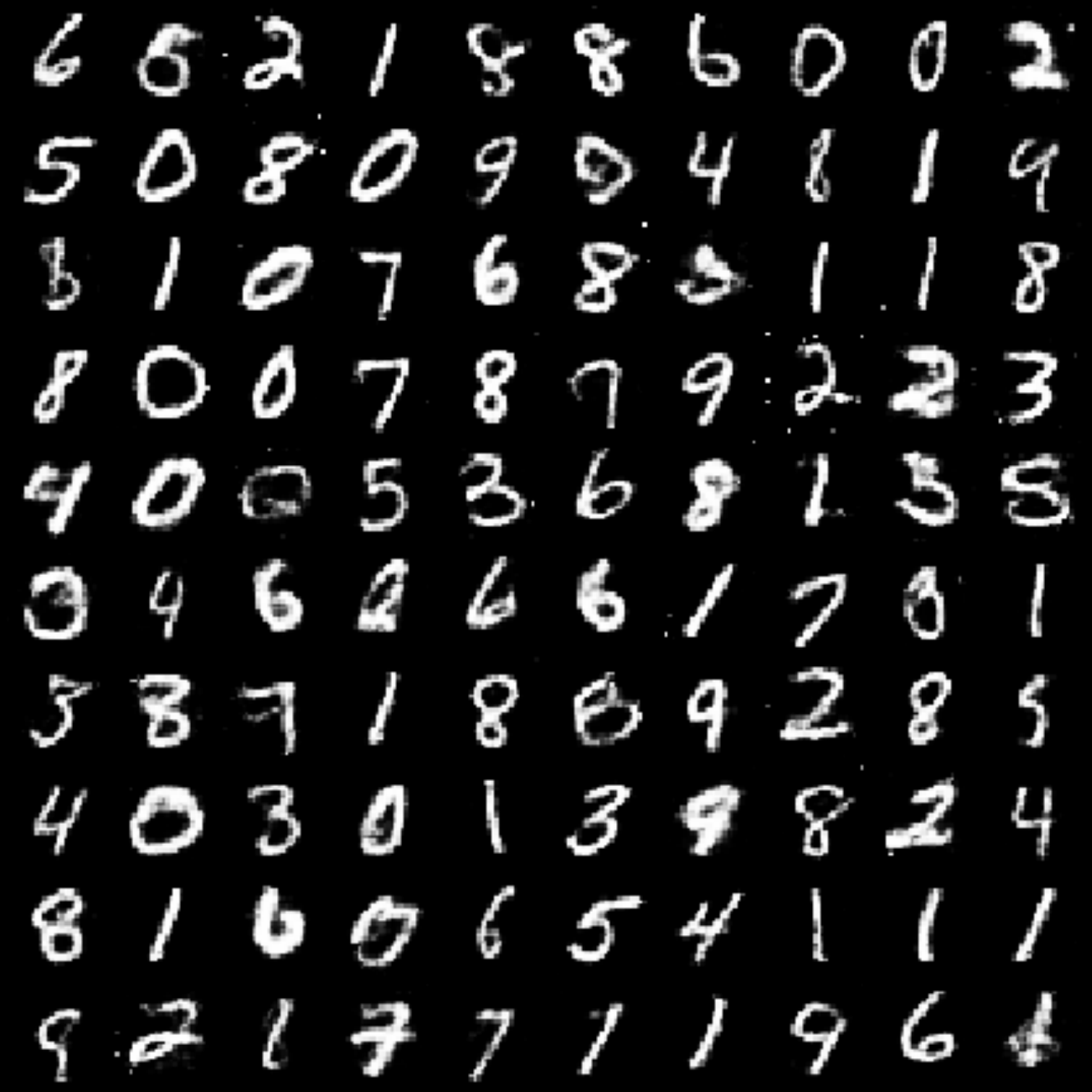}
         \caption{HGAN}
         \label{fig:three sin x}
     \end{subfigure}&
    \begin{subfigure}[b]{0.25\textwidth}
         \centering
         \includegraphics[width=\textwidth]{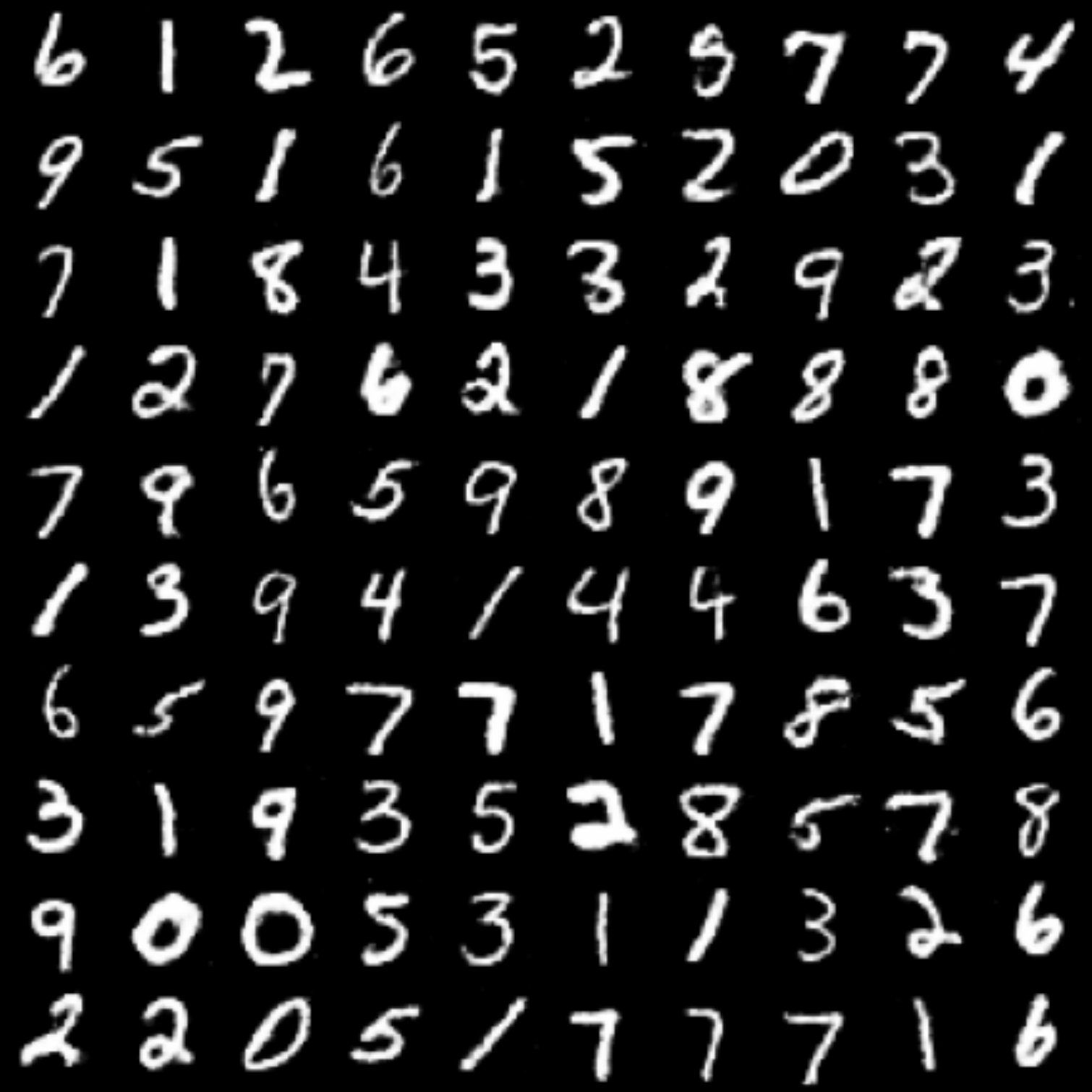}
         \caption{HCGAN}
         \label{fig:three sin x}
     \end{subfigure}&
    \begin{subfigure}[b]{0.25\textwidth}
         \centering
         \includegraphics[width=\textwidth]{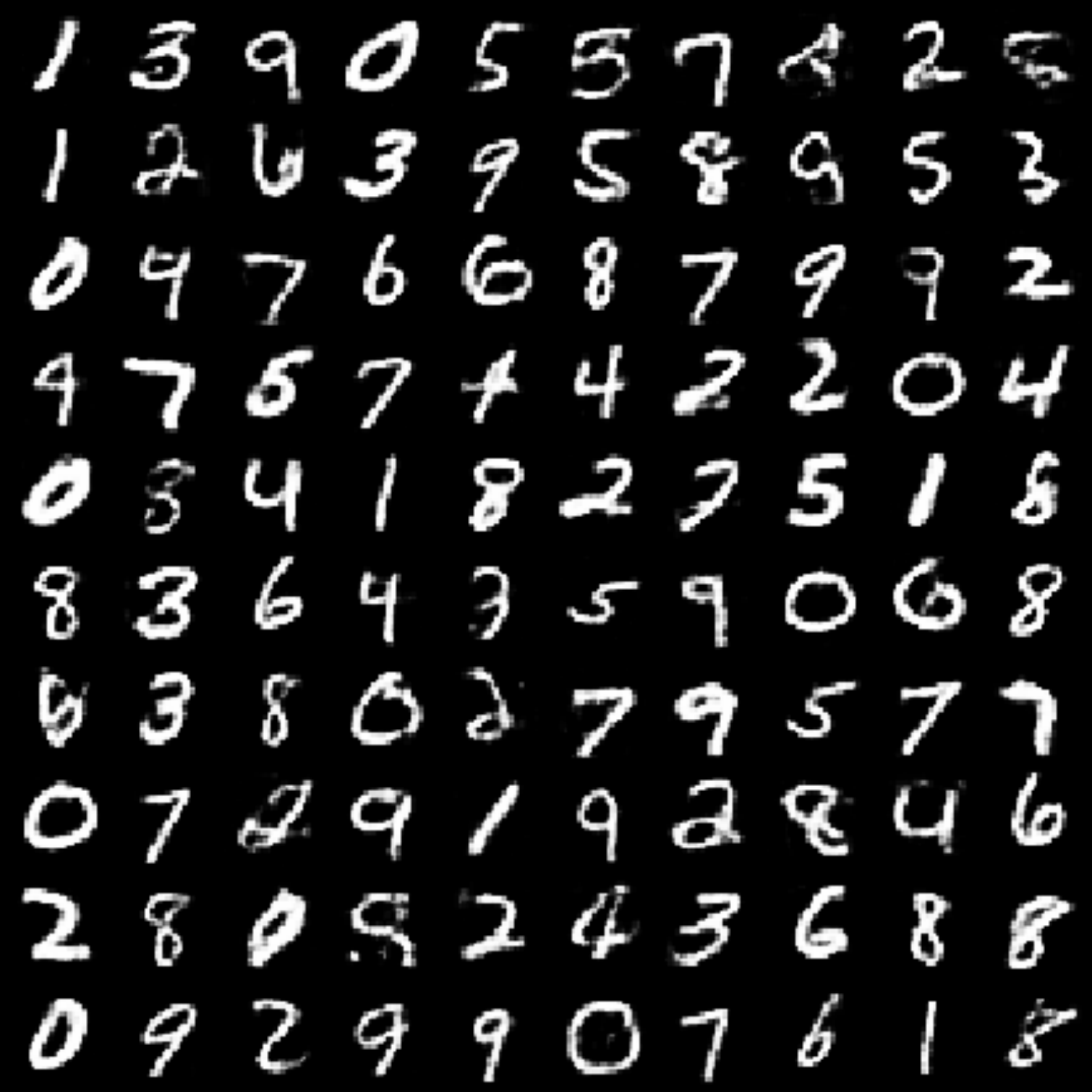}
         \caption{HWGAN}
         \label{fig:three sin x}
     \end{subfigure}
\end{tabular}
\caption{Images of the best results generated by each architecture, on MNIST and measured by the FID.}
\end{figure*} 
 

\begin{table}[H]
  \centering
  \begin{tabular}{|c|c|c|c|c|}
  \hline
  Architecture HGAN& $c_{d}$ &$c_{g}$& FID $\downarrow$ & IS $\uparrow$ \\
  \hline
  $\mathcal{D}_{eeee}\mathcal{G}_{eeee}$&-&-& 67.291 & 8.441\\
  \hline
  $\mathcal{D}_{heee}\mathcal{G}_{ehhe}$&1e-3&1e-3& 28.320 & 8.394\\
  \hline
  $\mathcal{D}_{ehhh}\mathcal{G}_{eehe}$&1e-5&1e-3&\textbf{18.697}&\textbf{9.291}\\
  \hline
  \end{tabular}
  \caption{The best HGAN results, measured by FID and IS with ten seeds for each configuration.}
  \label{table_hgan}
\end{table}

\begin{table}[H]
  \centering
  \begin{tabular}{|c|c|c|c|c|}
  \hline
  Architecture HWGAN& $c_{d}$ &$c_{g}$& FID $\downarrow$ & IS $\uparrow$ \\
  \hline
  $\mathcal{D}_{eeee}\mathcal{G}_{eeee}$&-&-& 16.130 & 9.284\\
  \hline
  $\mathcal{D}_{eeee}\mathcal{G}_{hhhe}$&-&0.1& 12.969&\textbf{9.424}\\
  \hline
  $\mathcal{D}_{eeee}\mathcal{G}_{ehhe}$&-&0.1&\textbf{12.884} & 9.291\\
  \hline
  \end{tabular}
  \caption{The best HWGAN results, measured by FID and IS.}
  \label{table_hwgan}
\end{table}

\begin{table}[H]
  \centering
  \begin{tabular}{|c|c|c|c|c|}
  \hline
  Architecture HCGAN& $c_{d}$ &$c_{g}$& FID $\downarrow$ & IS $\uparrow$ \\
  \hline
  $\mathcal{D}_{eeee}\mathcal{G}_{eeee}$&-&-& 11.588 & 9.910\\
  \hline
  $\mathcal{D}_{hhhh}\mathcal{G}_{hhee}$&0.01&0.01& 9.171 & 9.899\\
  \hline
  $\mathcal{D}_{hhee}\mathcal{G}_{hhee}$&0.01&0.01& \textbf{9.157} & 9.903\\
  \hline
    $\mathcal{D}_{hhhe}\mathcal{G}_{hhee}$&1e-5&1e-5& 10.940 & \textbf{9.919}\\
  \hline
  \end{tabular}
  \caption{The best HCGAN results, measured by FID and IS.}
  \label{table_hcgan}
\end{table}

\section{Conclusion}

This work shows that the HGAN, HCGAN, and HWGAN architectures can perform as well as or, in some cases, much better than the original euclidean architectures. The performance depends heavily on two main factors: the curvature through the $c$ parameter, and the architecture configuration. The best performance was achieved for the HGAN  with $c = 10^{-3}$, HCGAN with $c=10^{-2}$, and for the HWGAN with $c=10^{-1}$. For smaller values of $c$ the radius growths and the hyperbolic layers behave as euclidean layers, consequently there is no improvement over the euclidean version. And for $c<10^{-6}$ the networks did not converge because of numerical instability. One last consideration for the $c$ value was to make a distinction between the $c$ used in the discriminator that can be different from the $c$ used in the generator. This was because the different nature of the tasks performed by discriminator and generator, and it had as a consequence the best improvement of performance in the HGAN as showed in table \ref{table_hgan}.
For the configurations, the experiments show that the generator's $EHE$ and the $HE$ configurations had better performance. For the discriminator the $HE$ and the $EH$ configurations worked better. However, the HWGAN never showed a performance improvement with hyperbolic layers in the discriminator. When the discriminator was in the $EH$ configuration, the logarithmic map at the network output was not applied, this because it became unstable for one dimension.

\newpage
\flushend
\bibliographystyle{ieeetr}
\bibliography{references}

\section*{A. Visualization of experiments }

We implement a method to visualize all experiments for each architecture, HGAN, HCGAN, and HWGAN.\newpage

\begin{figure*}[h]
    \centering
    \includegraphics[scale=0.5]{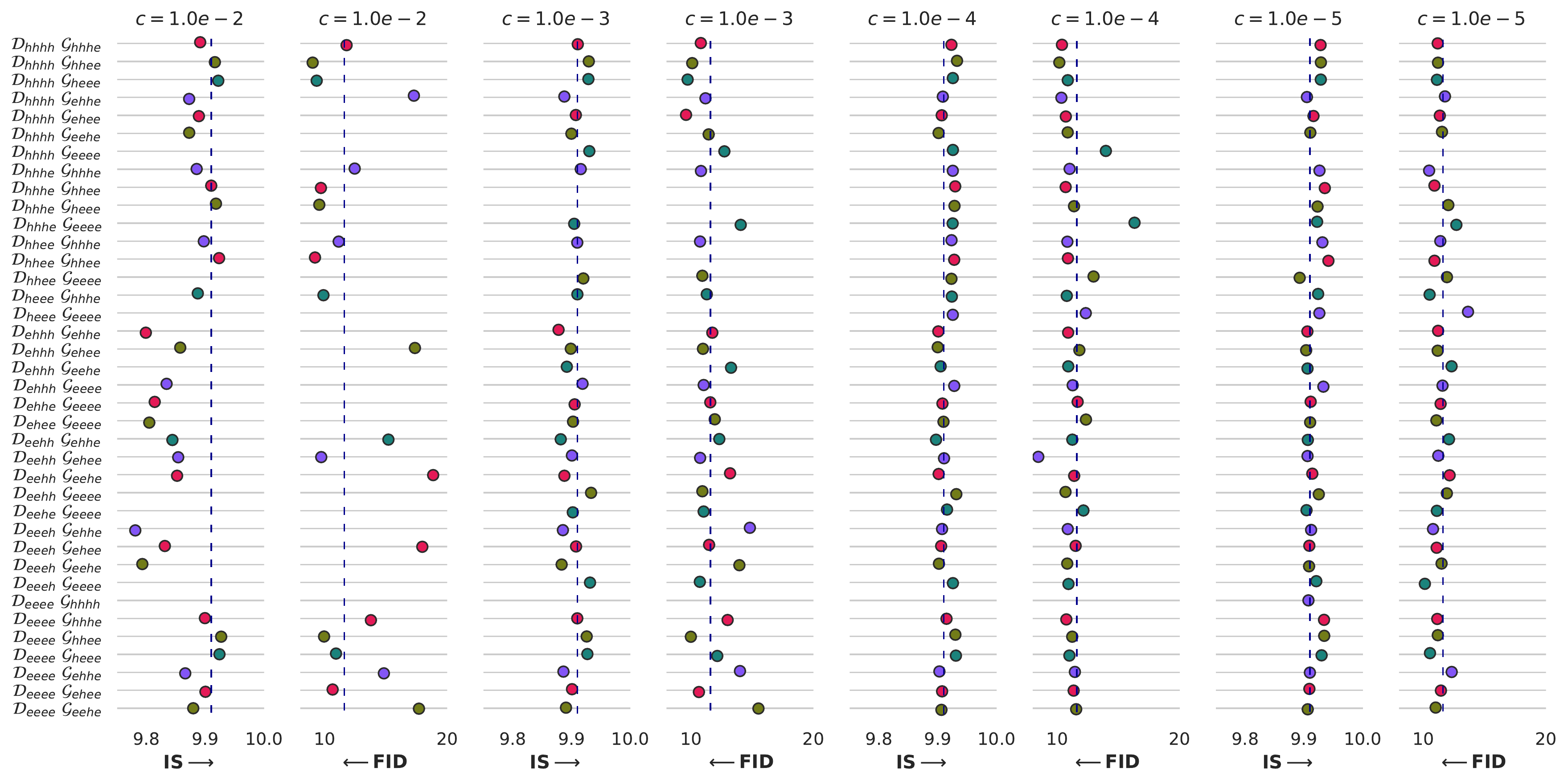}
    \caption{HCGAN experiments visualization with only hyperbolic discriminator or hyperbolic generator, and mixed hyperbolic configurations measure  with  FID and IS. The segment line indicate the euclidean CGAN performance.}
    \label{fig:hcgan_table}
\end{figure*}

\begin{figure*}[h]
    \centering
    \includegraphics[scale=0.5]{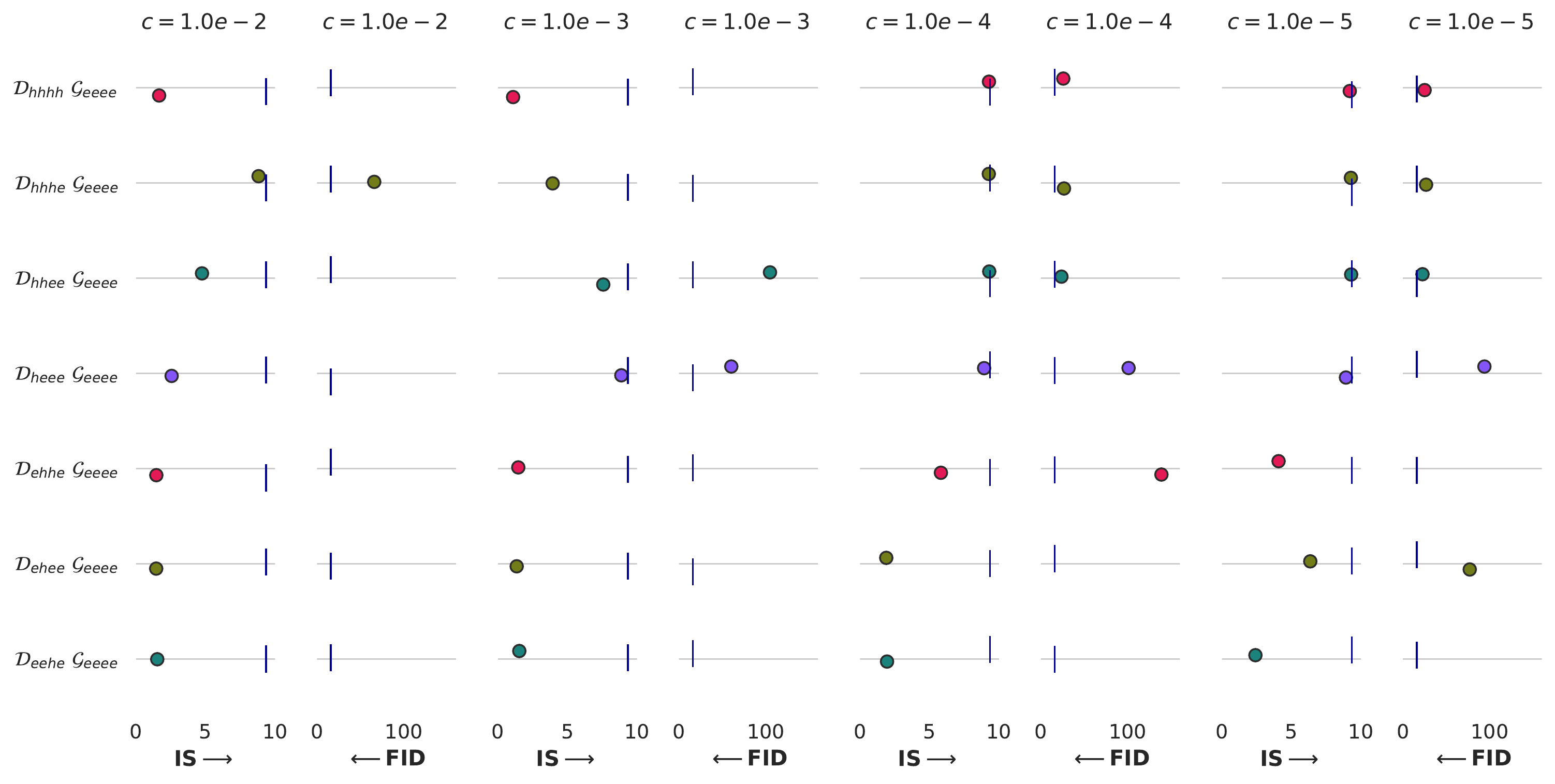}
    \caption{HWGAN experiments visualization measure  with  FID and IS. The segment line indicate the euclidean WGAN performance.}
    \label{fig:hwgan_table_d}
\end{figure*}

\begin{figure*}[h]
    \centering
    \includegraphics[scale=0.4, width=\textwidth]{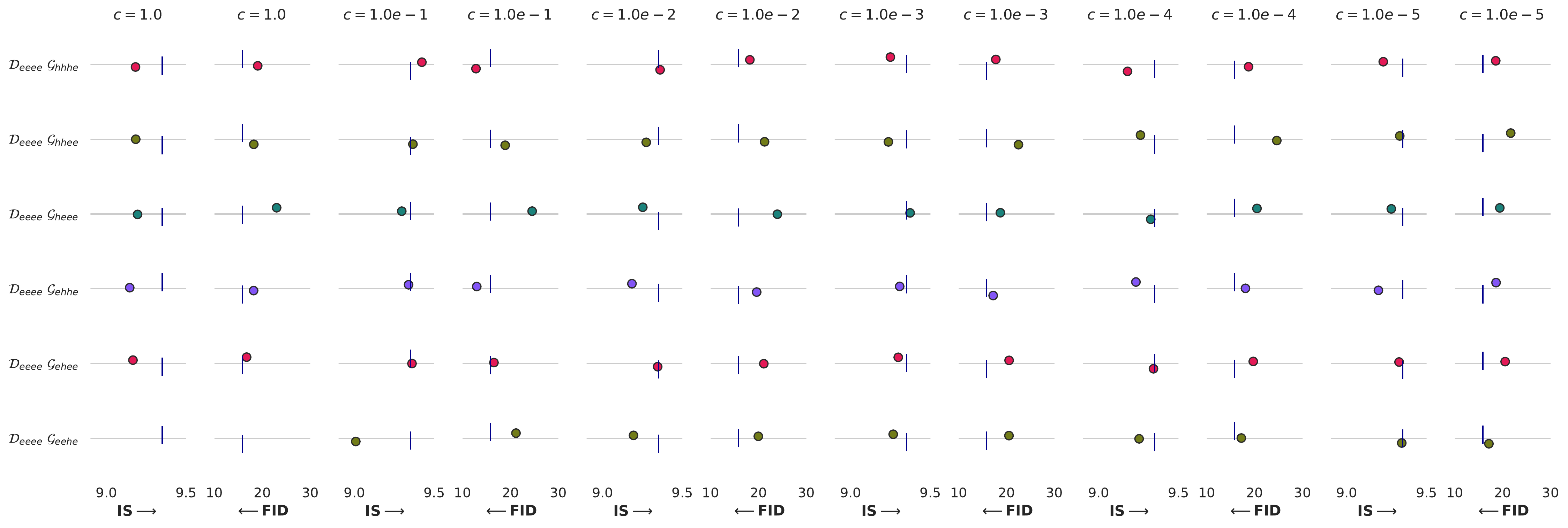}
    \caption{HWGAN experiments visualization with hyperbolic generator measure  with  FID and IS. The segment line indicate the euclidean WGAN performance.}
    \label{fig:hwgan_table_g}
\end{figure*}

Also, the visualization shows each configuration and c value. The rows show the configurations, for example $\mathcal{D}_{hhhh} \,\mathcal{G}_{eehh}$, and the columns indicate the c value. The IS and FID measurements are in the x-axis. The bigger values of IS represent better performance (showed by the $\rightarrow$ arrow), on the other hand smaller values on the FID score represent better values (showed by the $\leftarrow$ arrow). Therefore, when a network has good performance its IS and FID markers should be placed to the center of the column near each other, in contrast when the markers are far from each other and from the center it is an indicative of poor performance. Aditionally, the perfomance of the fully euclidean version is depicted by a vertical segment in each graph line. Figure \ref{fig:hcgan_table} shows the results for the HGAN, the best performing architectures can be seen for $c=10^{-3}$. The results for the HCGAN are displayed in figure \ref{fig:hcgan_table}, it is possible to observe that there are many configurations that presents better performance than the euclidean CGAN for all the tested values of $c$. Finally, figure \ref{fig:hwgan_table_d} and \ref{fig:hwgan_table_g} show the results obtained for the HWGAN. The best perfoming architecture was found for $c=0.1$.

\newpage

\end{document}